\documentclass{article}
\usepackage[utf8]{inputenc}
\usepackage{float}
\usepackage[preprint]{corl_2026}
\usepackage{amsmath,amssymb}
\usepackage{graphicx,booktabs,array,adjustbox,pifont}
\title{Watch, Recall, Act: Always-On Robots in Concurrent\\Embodied Streams}
\author{
  \textbf{Ding Yi\textsuperscript{1} \quad Peiwen Sun\textsuperscript{2} \quad Chenchu Rong\textsuperscript{3} \quad Jianan Wang\textsuperscript{4}} \\
  \textbf{Xili Dai\textsuperscript{5} \quad Xiangyu Yue\textsuperscript{2} \quad Xi Lin\textsuperscript{1}\thanks{Corresponding author: Xi Lin (\texttt{linxi234@sjtu.edu.cn}).}} \\
  \textsuperscript{1}Shanghai Jiaotong University \quad \textsuperscript{2}The Chinese University of Hong Kong \\
  \textsuperscript{3}Nanjing University \quad \textsuperscript{4}Astribot \quad \textsuperscript{5}Juxi Tech \\
  \texttt{onedonedone@sjtu.edu.cn} \quad \texttt{linxi234@sjtu.edu.cn}
}
\hypersetup{pdfauthor={Ding Yi, Peiwen Sun, Chenchu Rong, Jianan Wang, Xili Dai, Xiangyu Yue, Xi Lin}}

\begin{document}

\maketitle
\begin{abstract}
An always-on robot faces an endless stream that never resets: instructions arrive and lapse, the scene changes, and its own past actions reshape what it must reason about. Today's action models are built for the opposite: a fixed instruction, no mid-task intervention, single-step reasoning. In an open-ended world a robot must \emph{watch} a live stream for far-future cues, \emph{recall} its own far-past actions, and \emph{act} on them under dual-arm concurrency. We present \textbf{ARMS} (\textbf{A}lways-on \textbf{R}obot in \textbf{M}ulti-modal \textbf{S}treams), a deliberately simple streaming policy: a single pretrained $\pi _{0.5}$ backbone augmented by three lightweight modules that turn live perception, embodied states, and the robot's own past actions into context the backbone reads before it acts. The modules update this context \emph{asynchronously}, so watching and recalling never block acting and the two arms act at once. Rather than inventing new mechanisms, ARMS \emph{integrates} these learned context providers with an \emph{agent-causal} self-history that logs which arm did what, and when. To supervise them without extra annotation, we build \textbf{ARMS Dataset}, whose staged construction script itself labels every module from real dual-arm teleoperation. Trained on it, ARMS reaches 45\% on the combined task against 28\% for the strongest of our four main baselines, and ablations confirm the memory module, the embodied-state head, and asynchronous concurrency are each necessary.
\end{abstract}
% The introduction now starts below the abstract on page 1.

\section{Introduction}\label{sec:1}
Recent multimodal models are moving from offline frameworks toward \emph{streaming} understanding over an endless live stream \citep{ref1,ref2,ref3,ref4,ref5}. Such a model is at once \emph{proactive}, monitoring the stream and acting on a standing condition only when it is met, and \emph{retrospective}, recalling what happened far earlier in the stream. Either way, the information that decides an action is gathered far apart \emph{in time} from the moment to act.

For a robot, streaming is not optional. Unlike a video model that can watch a clip offline and then answer, an embodied agent lives in a single, never-reset timeline: its camera never closes, instructions arrive and lapse over hours, and it must perceive and act at once. Proactive watching and retrospective recall are thus prerequisites for long-horizon deployment, not add-ons. Action matters just as much to streaming in return: its proper output is a physical act, not an answer, and because a robot's own actions reshape the very scene it must interpret, what it did becomes part of what it must understand, something no passively observed stream demands.

% Place the teaser at the top of page 2, after the opening motivation.
\clearpage
\begin{figure}[!t]
\centering
\includegraphics[width=0.9\linewidth]{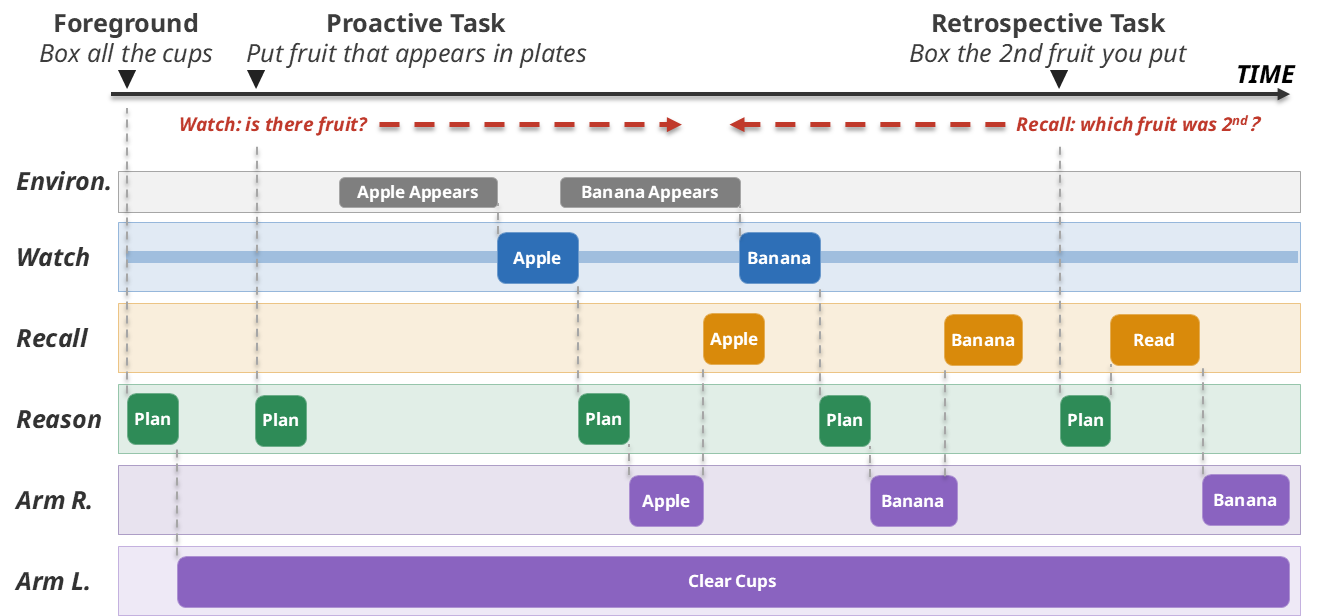}
\caption{\textbf{Timeline of a Streaming Action Task.} On the timeline of an always-on stream, different tasks may overlap: a \emph{foreground} task, a \emph{proactive} task, and a \emph{retrospective} query. Handling them requires the model to watch, recall, and act at the right moment with the right embodiment.}\label{fig:1}
\end{figure}

Three demands define this always-on regime (Figure \ref{fig:1}), which we abbreviate \emph{watch, recall, act}: the robot must \textbf{watch} a never-ending first-person stream for far-future triggers (``stow anything placed in front of you''), \textbf{recall} its own far-past actions to resolve self-referential commands (``the second one you stowed''), and \textbf{act} on a real dual-arm robot rather than answer a question about the video.

Vision-language-action models already supply these ingredients in isolation: streaming memory \citep{ref6,ref7}, mid-task switching \citep{ref8}, and always-on operation \citep{ref9}. None sustains their combination on a dual-arm robot, where three things must hold at once. A standing rule can fire long after it is set, so the robot must keep watching. A self-referential command can be answered only from a record of what the robot itself did, not from the current frame. And because two arms share and continually disturb one scene, acting safely means knowing when they are about to collide. The combination, not any one capability, is the hard part; and it is genuinely an \emph{action} problem, since replacing the arms with a multiple-choice answer makes the live scene, the foreground to sustain, and the self-perturbed grasp all disappear.

We make three contributions. \textbf{(1)} We present \textbf{ARMS} (Always-on Robot in Multi-modal Streams), a deliberately simple policy: a single pretrained $\pi _{0.5}$ backbone augmented by three lightweight modules that turn live perception, embodied state, and the robot's past actions into context the backbone reads before it acts (\S{}\ref{sec:3}); the contribution is this \emph{integration} and the \emph{agent-causal} self-history, not a new network or planner. \textbf{(2)} We build \textbf{ARMS Dataset}, a benchmark for this combination, from real dual-arm data whose staged construction itself supplies the supervision each module needs (\S{}\ref{sec:3.7}). \textbf{(3)} We deploy ARMS on a real dual-arm robot and, through a Probe/Online comparison, localize the difficulty in execution rather than perception (\S{}\ref{sec:4}). On ARMS Dataset and a real dual-arm robot, ARMS reaches $\sim$45\% versus $\sim$28\% for the best baseline in Table~\ref{tab:2}, and ablations confirm that the memory module, the embodied-state head, and asynchronous concurrency are each necessary.

\begin{table}[!t]
\centering
\caption{\textbf{Capability comparison on the always-on, dual-arm regime.} \emph{Always-on}: runs continuously without a per-task reset; \emph{Proactive}: acts on a standing condition that may fire long after it is set; \emph{Retrospective}: draws on a persistent memory of its own past; \emph{Interruptible}: can switch to a newly arriving instruction mid-execution; \emph{Concurrent}: runs the inserted action on a free arm \emph{without halting} the foreground. Prior systems each cover only part of this regime: several are always-on, can be interrupted, or keep a memory, but none is proactive \emph{and} runs the inserted action concurrently rather than switching away from the foreground; ARMS covers all five axes of the regime it targets (which prior systems were not designed for).}\label{tab:1}
\small
\setlength{\tabcolsep}{0pt}
\begin{adjustbox}{max width=\linewidth}
\begin{tabular*}{308.642bp}{@{\extracolsep{\fill}}lccccc@{}}
\toprule
Method & Always-on & Proactive & Retrospective & Interruptible & Concurrent \\
\midrule
SwitchVLA \citep{ref8} & \textcolor[rgb]{0.7019607843137254,0.0,0.0}{\ding{55}} & \textcolor[rgb]{0.7019607843137254,0.0,0.0}{\ding{55}} & \textcolor[rgb]{0.7019607843137254,0.0,0.0}{\ding{55}} & \textcolor[rgb]{0.0,0.6,0.0}{\ding{51}} & \textcolor[rgb]{0.7019607843137254,0.0,0.0}{\ding{55}} \\
MemoryVLA \citep{ref6} & \textcolor[rgb]{0.7019607843137254,0.0,0.0}{\ding{55}} & \textcolor[rgb]{0.7019607843137254,0.0,0.0}{\ding{55}} & \textcolor[rgb]{0.0,0.6,0.0}{\ding{51}} & \textcolor[rgb]{0.7019607843137254,0.0,0.0}{\ding{55}} & \textcolor[rgb]{0.7019607843137254,0.0,0.0}{\ding{55}} \\
VITA-E \citep{ref10} & \textcolor[rgb]{0.0,0.6,0.0}{\ding{51}} & \textcolor[rgb]{0.7019607843137254,0.0,0.0}{\ding{55}} & \textcolor[rgb]{0.7019607843137254,0.0,0.0}{\ding{55}} & \textcolor[rgb]{0.0,0.6,0.0}{\ding{51}} & \textcolor[rgb]{0.7019607843137254,0.0,0.0}{\ding{55}} \\
Habilis-$\beta$ \citep{ref9} & \textcolor[rgb]{0.0,0.6,0.0}{\ding{51}} & \textcolor[rgb]{0.7019607843137254,0.0,0.0}{\ding{55}} & \textcolor[rgb]{0.7019607843137254,0.0,0.0}{\ding{55}} & \textcolor[rgb]{0.7019607843137254,0.0,0.0}{\ding{55}} & \textcolor[rgb]{0.7019607843137254,0.0,0.0}{\ding{55}} \\
RoboStream \citep{ref11} & \textcolor[rgb]{0.0,0.6,0.0}{\ding{51}} & \textcolor[rgb]{0.7019607843137254,0.0,0.0}{\ding{55}} & \textcolor[rgb]{0.0,0.6,0.0}{\ding{51}} & \textcolor[rgb]{0.7019607843137254,0.0,0.0}{\ding{55}} & \textcolor[rgb]{0.7019607843137254,0.0,0.0}{\ding{55}} \\
\midrule
\textbf{ARMS (ours)} & \textcolor[rgb]{0.0,0.6,0.0}{\ding{51}} & \textcolor[rgb]{0.0,0.6,0.0}{\ding{51}} & \textcolor[rgb]{0.0,0.6,0.0}{\ding{51}} & \textcolor[rgb]{0.0,0.6,0.0}{\ding{51}} & \textcolor[rgb]{0.0,0.6,0.0}{\ding{51}} \\
\bottomrule
\end{tabular*}
\end{adjustbox}
\end{table}

\section{Related Work}\label{sec:2}
Table \ref{tab:1} contrasts ARMS with recent systems, each of which addresses only part of the always-on, dual-arm regime; we discuss the axes below.

\textbf{Robot with Temporal Reasoning.} A growing line equips VLAs with memory for long-horizon, history-dependent control \citep{ref12,ref13,ref14,ref15}: MemoryVLA buffers perceptual and cognitive tokens \citep{ref6}, ReMem-VLA uses dual-level recurrent queries \citep{ref7}, RoboMME standardizes memory evaluation \citep{ref16}, and, closest to us, RoboStream tracks state transitions and object permanence with a causal spatio-temporal graph \citep{ref11}; History-Dependent-Manipulation grounds language to previously manipulated, even occluded, objects \citep{ref17}. Yet all such memory is \emph{passive} or \emph{world-state-centric}. ARMS instead keeps an \emph{agent-causal} self-history logging \emph{which arm} performed \emph{which} guard-driven action, and when, queryable by language on an always-on robot: recording the acting arm keeps two hands' concurrent records from colliding and lets ``the one the left arm stowed'' resolve.

\textbf{Streaming, proactive, and interruptible control.} On the perception side, proactive video LLMs decide \emph{when} to respond on a live stream \citep{ref18,ref19,ref20,ref21}: ProactiveVideoQA contributes a timeliness metric \citep{ref2}, OmniPro the Probe/Online protocol we adopt \citep{ref3}, Em-Garde a propose--match scheme \citep{ref5}, and RIVER mirrors our retrospective/live/proactive framing \citep{ref4}. Yet all \emph{answer}: the output is recognition, never a physical act, which is where ARMS differs. We inherit Dispider's non-blocking asynchronous skeleton \citep{ref1} and replace its blocking reaction track with embodied dual-arm action; lacking a foreground task, physical action, or self-history, Dispider never faces ARMS's conflicts. On the control side, SwitchVLA switches tasks mid-execution \citep{ref8}, VITA-E interleaves seeing, hearing, speaking, and acting via a dual active/standby model \citep{ref10}, and Habilis-$\beta$ targets reset-free operation \citep{ref9}; recent dual-system and real-time policies push streaming action further \citep{ref22,ref23,ref24,ref25}. None runs the foreground arm in parallel with an inserted action, commits it to a queryable self-history, or arbitrates more demands than free hands: the dual-arm, concurrent, overload conflicts at ARMS's center.

\textbf{Robot Foundation Models.} ARMS builds on the growing family of general robot foundation models \citep{ref26,ref27,ref28,ref29,ref30} rather than competing with them: its action expert is fine-tuned from $\pi _{0.5}$, whose pretraining and language-subtask conditioning we build on \citep{ref31}. Complementary foundations ARMS does \emph{not} require but could draw on include spatially grounded encoders for relative references \citep{ref32,ref33} and discrete skill-tokenization vocabularies \citep{ref34,ref35}. Our contribution is orthogonal: the always-on, dual-arm, self-history layer that turns such a policy into an agent which acts on, records, and is later queried about its own behaviour.

\section{ARMS}\label{sec:3}
We present ARMS, a deliberately simple policy for the always-on, dual-arm setting. Rather than building a dedicated mechanism for each new demand, ARMS keeps a single pretrained backbone and turns everything else into \emph{context}: three lightweight modules each convert one signal (live perception, embodied state, and past actions) into tokens the backbone reads before it acts.

\begin{figure}[!t]
\centering
\includegraphics[width=\linewidth]{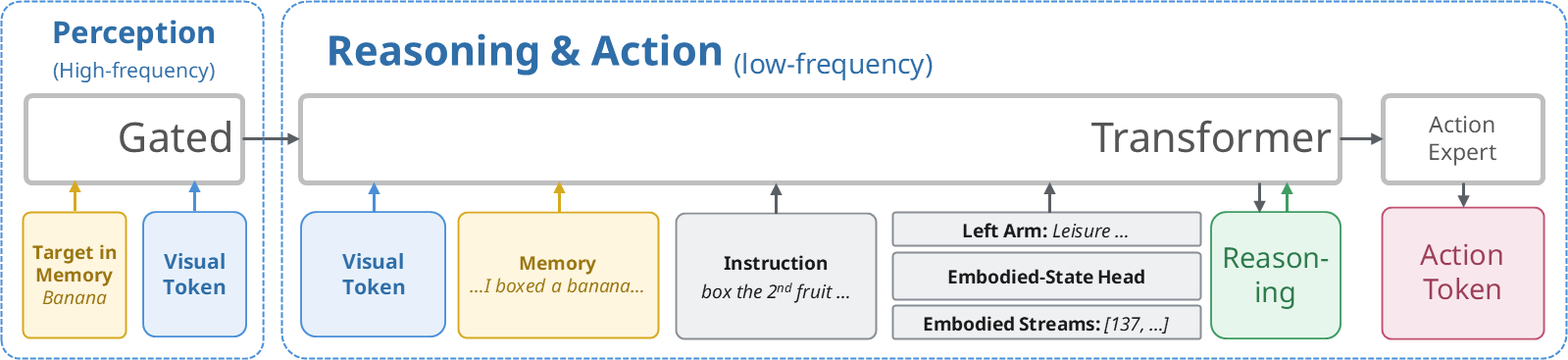}
\caption{\textbf{The ARMS architecture.} A high-frequency \emph{perception} path runs a gated visual trigger (does the guard's target object, e.g. the banana held in memory, appear?) and only on a positive trigger wakes the low-frequency \emph{reasoning-and-action} path: a single $\pi _{0.5}$ backbone (Transformer $+$ action expert) that reads the injected context (visual tokens, compressed memory, the language instruction, and the embodied-state head's per-arm primitives and streams) and autoregresses the dual-arm action token. The three modules share one backbone and one context interface rather than acting as separate planners.}\label{fig:2}
\end{figure}

\subsection{Problem setup}\label{sec:3.1}
A robot runs on a single, never-reset first-person video stream while controlling two arms. At each step $t$ it receives an observation $o_{t}$ (RGB with proprioception) and, for each arm, continues, begins, or withholds an action; the episode never terminates, and the scene is continually altered by the robot's own two arms. Two kinds of language input drive it. A \emph{guard} is a standing rule (``whenever an item appears, stow it'') that stays active and may fire long after it was set (the \emph{proactive} face). A \emph{recall} instruction refers to the robot's own past actions (``put the second one you stowed in the box''), answerable only from a record of what the robot did, not from $o_{t}$ (the \emph{retrospective} face). The deciding information is thus decoupled in time from the moment to act; and because two arms share and disturb one scene, acting also requires knowing when the arms are about to collide. ARMS must \emph{watch}, \emph{recall}, and \emph{act} under exactly this combination.

\subsection{Overview: one backbone, three context providers}\label{sec:3.2}
The backbone is a pretrained $\pi _{0.5}$ vision-language-action model \citep{ref31} (a vision-language transformer with an action expert), reused with its original architecture, including its vision encoder. Around it, ARMS adds three light modules (Fig. \ref{fig:2}): a \textbf{visual perception} path reusing $\pi _{0.5}$'s vision head as a streaming, Dispider-style trigger gate (\S{}\ref{sec:3.3}); an \textbf{embodied-state head} reporting, in one sentence, each arm's current action primitive and whether the two arms are on a colliding path (\S{}\ref{sec:3.4}); and a \textbf{memory module} compressing past actions and re-injecting the relevant ones for recall (\S{}\ref{sec:3.5}). Each emits tokens (visual tokens or a short sentence) appended to $\pi _{0.5}$'s context, over which the backbone autoregresses the dual-arm action. The modules update this context \emph{asynchronously} while the heavier action forward reads it on demand, so perception and memory continue even while the arms execute, non-blocking, following Dispider \citep{ref1}; this is what makes task-level \emph{concurrency} possible. The perception gate runs at \emph{high frequency} to keep watching the stream, while the heavier backbone reasons and acts at a \emph{lower frequency}, woken only when a trigger fires. Each module is small and purpose-built, yet the three share one backbone and one interface, its context, with no separate planner, scheduler, or arbiter: watching, recalling, and collision-safe acting all reduce to what context is placed in front of $\pi _{0.5}$. Because each module's output is explicit, it can also be read \emph{without} executing, giving the recognition probe of \S{}\ref{sec:4}.

% Queue the qualitative figure for the top of page 5.
\begin{figure}[!t]
\centering
\includegraphics[width=\linewidth]{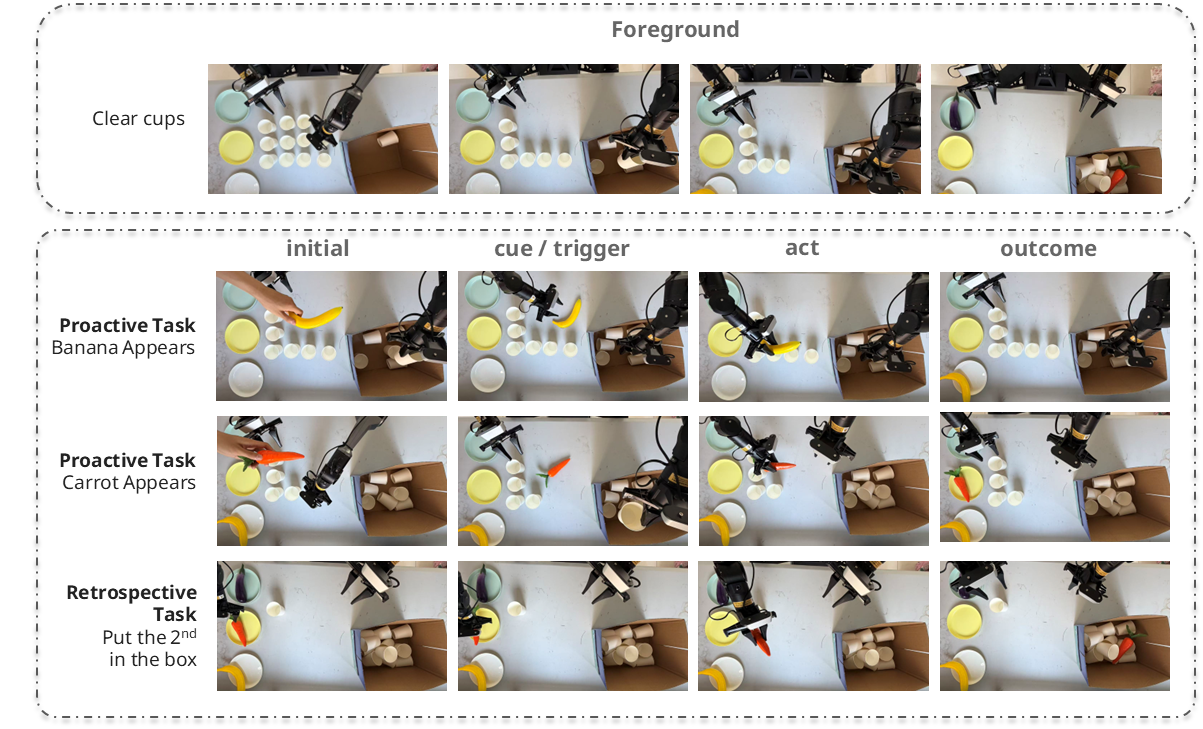}
\caption{\textbf{Qualitative results on the real robot,} a single never-reset concurrent episode. \emph{Foreground} row: clearing cups into the box. Below (columns \emph{initial/cue/act/outcome}): two \emph{proactive} guards fire as a banana then a carrot appear and a free arm stows each; the \emph{retrospective} ``put the second one you stowed in the box'' resolves to the carrot and boxes it under inter-arm occlusion.}\label{fig:3}
\end{figure}

\subsection{Visual perception: a streaming trigger gate}\label{sec:3.3}
\begingroup\looseness=-1 ARMS adds no new perception network: the live first-person stream is encoded by $\pi _{0.5}$'s own vision head. Following Dispider \citep{ref1}, we decouple this visual path from the heavy action forward and run it continuously as a lightweight, non-blocking \emph{trigger gate}. Against each active guard, the gate makes a \emph{binary} decision on the current observation: whether the guard's target object is present and its condition holds (fire) or not (keep watching). While the answer is \emph{no}, the gate keeps scanning at full rate without waking the action expert, so an always-on stream stays cheap; the moment it turns \emph{yes}, the encoded visual tokens enter $\pi _{0.5}$'s context and the action expert is invoked to act. This gate is exactly the \emph{watch} operator: it is what lets a standing rule fire long after it was set, and what keeps the foreground undisturbed until a trigger actually occurs. Being a forward of the reused vision head, it adds near-zero parameters and keeps its tokens in a space the backbone already understands.\par\endgroup

\subsection{Embodied-state head: action primitive and collision in one sentence}\label{sec:3.4}
A small \textbf{embodied-state head} reads the current observation and proprioception and appends one short sentence to the context, carrying two things: each arm's current \emph{action primitive} (grasping, releasing, reaching, idle), giving the backbone an explicit handle on what it is doing (exactly what a recall query or a newly fired guard must reason against), and a \emph{collision check}, whether the two arms are on a colliding path given their current and intended motions. Phrasing both as one sentence (e.g. ``left arm is stowing a cup; right arm reaching the plate is clear'') lets the backbone, rather than a separate geometric planner, decide how to proceed: dispatch a concurrent action to a free arm when the path is clear, or serialize when it is not. Collision avoidance and arm coordination thus reduce to a sentence the backbone reads, not a module that overrides it. Physical execution additionally uses joint velocity limits and an operator-controlled emergency stop.

% Queue the main results table for the top of page 6.
\begin{table}[!t]
\centering
\caption{\textbf{Main results and ablations on ARMS Dataset (Online protocol).} Percentages. Recall $=$ cross-interruption recall accuracy; FG-Res. $=$ foreground-continuation rate; Hist. $=$ self-history consistency; Concur. $=$ dual-arm non-interference; Overload $=$ priority-ordering accuracy; PAUC $=$ proactive-timeliness; EAS $=$ combined-task action success (headline). Oracle/Human is an expert teleoperator, an empirical upper bound.}\label{tab:2}
\small
\setlength{\tabcolsep}{0pt}
\begin{adjustbox}{max width=\linewidth}
\begin{tabular*}{378.683bp}{@{\extracolsep{\fill}}lccccccc@{}}
\toprule
Method & Recall & FG-Res. & Hist. & Concur. & Overload & PAUC & \textbf{EAS} \\
\midrule
Fixed-rate VLA (sliding window) & 10 & 22 & 5 & 30 & 9 & 20 & 16 \\
MemoryVLA \citep{ref6} & 44 & 28 & 40 & 27 & 12 & 30 & 24 \\
SwitchVLA \citep{ref8} & 15 & 58 & 18 & 50 & 20 & 44 & 28 \\
Recognition$\rightarrow$Controller & 40 & 21 & 45 & 29 & 15 & 40 & 19 \\
\midrule
\textbf{ARMS (ours)} & \textbf{72} & \textbf{76} & \textbf{82} & \textbf{78} & \textbf{71} & \textbf{68} & \textbf{45} \\
w/o memory module & 31 & 74 & 27 & 77 & 69 & 66 & 32 \\
w/o agent-causal tags & 50 & 75 & 55 & 77 & 69 & 66 & 38 \\
memory w/o temporal order & 48 & 75 & 58 & 78 & 70 & 67 & 36 \\
w/o embodied-state head & 58 & 58 & 81 & 68 & 70 & 66 & 34 \\
hard interrupt (no async concurrency) & 68 & 41 & 80 & 38 & 55 & 64 & 30 \\
\midrule
Oracle / Human & 93 & 92 & 96 & 90 & 88 & 88 & 85 \\
\bottomrule
\end{tabular*}
\end{adjustbox}
\end{table}

\subsection{Self-action memory by compression}\label{sec:3.5}
To answer recall, ARMS keeps a memory $\mathcal{H}$ of its own past actions and re-injects the relevant part into the context. Following MemoryVLA \citep{ref6}, $\mathcal{H}$ is \emph{compressed} rather than a growing log: each completed action is summarized into a compact entry (what was done, with which arm, to which object, and when), and entries are consolidated so memory and retrieval stay bounded on an unbounded stream. When a recall instruction arrives, the module retrieves the matching past actions and writes them back as a short sentence (``earlier the right arm stowed the carrot second, the banana first'') so the backbone resolves ``the second one you stowed'' from what it actually did. This is also what tells two \emph{same-class} objects apart: two identical bananas are distinguished by \emph{when and how} each was handled, recorded in $\mathcal{H}$, not by appearance. At recall time, objects are matched to memory entries using stored object IDs under a last-known-identity assumption. Because recall concerns past actions absent from $o_{t}$, a memory-free policy can only guess from the current scene \citep{ref17}; compression is what keeps the retrospective face affordable over long horizons (\S{}\ref{sec:4}).

\subsection{Training}\label{sec:3.6}
ARMS is not trained from scratch. The $\pi _{0.5}$ backbone, including its vision head, is initialized from released pretrained weights and fine-tuned for the dual-arm setting, inheriting its pretraining. The three added modules are light and trained on ARMS Dataset (\S{}\ref{sec:3.7}): the embodied-state head from dataset-derived action-primitive and collision labels, the memory module from the recorded action history and instruction--target pairs of the same construction. Because this supervision comes from the data rather than from rollouts of the backbone, the modules train without a circular dependency on the policy they augment. ARMS is a single model and runs inference on a single 80 GB accelerator.

\subsection{ARMS Dataset}\label{sec:3.7}
To train and evaluate ARMS under this combination, we build \textbf{ARMS Dataset}, a benchmark that requires always-on streaming, dual-arm concurrency, agent-causal self-history, end-to-end action, and overload arbitration \emph{together}. Its core is data we collect ourselves on a \textbf{Cobot Magic} dual-arm platform \citep{ref36} (a low-cost Mobile-ALOHA-style rig \citep{ref37}, teleoperated): about $\sim$200 never-reset \emph{staged streams} (minutes each: a person places objects on cue while both arms work and a third demand forces deferral) and $\sim$1,200 short \emph{skill demonstrations} over $\sim$6 dual-arm skills and $\sim$15 object classes, with $\geq$2 same-class distractor pairs (e.g. two bananas). Controlling our own collection is what unlocks the regime (triggering, concurrency, and overload are enacted on the real robot, not approximated in simulation), and every scored action is a real teleoperated demonstration, never synthetic; we optionally add LIBERO \citep{ref38} only to scale up. Because we \emph{construct} the streams, the build script is itself an oracle: each module is supervised by a label the script already wrote when it staged the situation. We will publicly release ARMS Dataset and the construction code.

% Continue naturally after the added identity and safety clarifications.

The primary \textbf{Online} protocol scores end-to-end action over four families: a guard firing concurrently with a foreground task (A), recalling a past self-action by language (B), relocating a self-displaced target (C), and arbitrating when both arms are busy and a new guard fires (D), all under six metrics (cross-interruption recall, foreground-continuation, self-history consistency, dual-arm non-interference, overload-ordering, and a PAUC-style timeliness). A complementary \textbf{Probe} protocol, which we adopt from OmniPro \citep{ref3} (not a contribution of ours), scores recognition alone (whether the agent can tell \emph{when to act on what} without moving), so contrasting the two exposes a perception--action \emph{dissociation}. Evaluation is on the real Cobot Magic robot ($\sim$150 trials per method) with LIBERO as an optional scale-up.

\section{Experiments}\label{sec:4}
Our experiments answer three questions. \textbf{(Q1)} Does the combination ARMS targets open a measurable gap that current policies fall into, and is that gap one of \emph{action} rather than perception (\S{}\ref{sec:4.2})? \textbf{(Q2)} Is each added module necessary (\S{}\ref{sec:4.3})? \textbf{(Q3)} Does ARMS hold up on a real dual-arm robot (\S{}\ref{sec:4.4})?

\subsection{Setup}\label{sec:4.1}
\textbf{Baselines.} Our main comparison includes four policies, each strong on one ingredient but missing the combination: a \emph{Fixed-rate VLA} runs the $\pi _{0.5}$ backbone in a sliding window with no streaming context (``always acting''); \emph{MemoryVLA} \citep{ref6} adds memory but neither sustains a foreground task through an insertion nor records \emph{which arm} acted; \emph{SwitchVLA} \citep{ref8} switches tasks mid-execution but keeps no queryable self-history; and \emph{Recognition}$\rightarrow$\emph{Controller} drives a naive executor from the strongest Probe front-end, with no memory or embodied-state module, the ``answers but cannot do'' control. All four baselines use the same pretrained $\pi_{0.5}$ initialization, fine-tuning data, and backbone training budget as ARMS; ARMS additionally trains its context modules. We evaluate the default \textbf{ARMS} ($\sim$5B) and, in \S{}\ref{sec:4.3}, its ablations.

\textbf{Protocol and metrics.} Evaluation is \emph{closed-loop}: we recompose recorded scenes into always-on streams and roll the policy out on the real robot (the live $N=60$ runs of \S{}\ref{sec:4.4} skip recomposition), so every physical metric reflects \emph{executed} actions, not open-loop matching. We use a 90/10 train/validation split over complete staged streams, with held-out paraphrases for recall evaluation. Beyond the six per-family metrics (\S{}\ref{sec:3.7}) we report \emph{combined-task action success} (EAS): an episode succeeds only if, in one uninterrupted run, the foreground completes, every fired guard is served on a free arm without collision, recall returns the correct past object, and overload ordering matches the priority labels. Overall EAS is weighted by evaluation counts across task families, whereas diagnostic metrics are computed on their respective applicable subsets. EAS is the headline; the six metrics localize \emph{where} a policy breaks. All percentages.

\subsection{The combination gap}\label{sec:4.2}
Table \ref{tab:2} (top) gives the main result: each baseline occupies the corner of its one strength and vacates the rest. MemoryVLA recalls but its foreground collapses under interruption (FG-Res. 28); SwitchVLA continues the foreground but cannot retrieve a past self-action (Recall 15); the Fixed-rate VLA and Recognition$\rightarrow$Controller sit low on both. The high-recall, high-continuation region is empty but for the oracle and ARMS (72/76). End to end, ARMS reaches 45\% EAS versus 28\% for SwitchVLA, the strongest baseline in Table~\ref{tab:2} ($+17$ percentage points), still $\sim$40 points below the $\geq$85\% human/oracle ceiling. A Probe/Online comparison places the gap in \emph{execution}, not perception: recognition stays high ($>$80\%) while action success decays under concurrency and self-perturbation, so this is an \emph{action} problem, not video QA.

\subsection{Ablations}\label{sec:4.3}
Table \ref{tab:2} (middle) removes one module at a time; each collapses on the metric it serves, and on EAS. Removing the \emph{memory module} collapses recall ($72 \rightarrow 31$): without provenance the policy can neither tell same-class items apart nor order them, falling below single-category chance and dragging history consistency down. Stripping only the \emph{temporal order} keeps category recognition but destroys ordinal recall (``the \emph{second} one''), so recall falls without collapsing (EAS $36$). Removing the \emph{embodied-state head} hurts foreground continuation and non-interference ($76 \rightarrow 58$, $78 \rightarrow 68$), as the backbone loses its handle on when the arms interfere. Replacing asynchronous concurrency with a \emph{hard interrupt} is sharpest, cratering both ($76 \rightarrow 41$, $78 \rightarrow 38$): acting on a free arm beats interrupting. The visual path is the backbone's own vision head, not separately removable. Removing the memory module, the embodied-state head, or asynchronous concurrency drops EAS to 32, 34, and 30 respectively, each a large fall from 45, so each is necessary. Keeping the same memory capacity but dropping the \emph{agent-causal} which-arm/when labels (a world-state memory) collapses recall and history ($72 \rightarrow 50$, $82 \rightarrow 55$), so the gain is the provenance labelling, not memory size.

% Queue the real-robot table for the top of page 8.
\begin{table}[!t]
\centering
\caption{\textbf{Real dual-arm robot} (three rounds of randomized task arrangements, 20 trials per round; $N=60$ per method/segment). Percentages are pooled over rounds and rounded to the nearest integer. ``Best baseline'' is the strongest competitor on each segment.}\label{tab:3}
\small
\setlength{\tabcolsep}{0pt}
\begin{adjustbox}{max width=\linewidth}
\begin{tabular*}{376.130bp}{@{\extracolsep{\fill}}llcc@{}}
\toprule
Segment & Metric & ARMS & Best baseline \\
\midrule
Watch $+$ concurrency (stow fruit while clearing cups) & success / FG-resume & 65 / 70 & 20 / 28 \\
Recall (``the second one you stowed'') & success & 60 & 13 \\
Self-perturbation (inter-arm occlusion) & relocation success & 62 & 18 \\
Long stream ($\geq$10 min, no reset) & retention & 67 & 30 \\
\bottomrule
\end{tabular*}
\end{adjustbox}
\end{table}

\subsection{Real-robot deployment and qualitative results}\label{sec:4.4}
We deploy ARMS on a real Cobot Magic robot for the full episode of \S{}\ref{sec:3.7} (Figure \ref{fig:3}, Table \ref{tab:3}; $N=60$/segment). While clearing cups into a box, a free arm stows a banana then a carrot as each appears (concurrent execution under frequent inter-arm occlusion); then ``put the second one you stowed in the box'' makes ARMS recall the carrot from $\mathcal{H}$ and box it, where baselines drop the foreground or fetch the wrong object. \emph{Honest scope}: this demo uses two \emph{distinct} items, so recall could lean on appearance; the same-class case (identical objects, separated only by provenance and order) and full overload are quantified on ARMS Dataset (families B, D), and large-scale real-robot evaluation is future work.

Over $\geq$10-min never-reset streams ARMS holds steady ($\mathcal{H}$ stays bounded), retaining $\sim$67\% vs $\sim$30\% (Table \ref{tab:3}, last row).

In sum, ARMS sustains proactive concurrency, self-history recall, and priority arbitration together (composed streams, real robot, long horizons) where single-ingredient policies do not, each module necessary. On one 80 GB accelerator the asynchronous modules keep median trigger-to-decision latency under 200 ms while both arms execute.

\section{Conclusion and Limitations}\label{sec:5}
We cast always-on dual-arm manipulation as acting at moments decoupled in time from the deciding information, a regime that is hard only when streaming memory, mid-task concurrency, and always-on operation must hold at once. \textbf{ARMS} meets it with a deliberately simple design: a single $\pi _{0.5}$ backbone integrating three lightweight context modules (visual perception, an embodied-state head, and a compressed agent-causal memory) through one shared context interface; the contribution is this integration, the agent-causal self-history, and a dataset whose construction supplies each module's supervision. On ARMS Dataset and a real dual-arm robot, ARMS beats policies built for any single ingredient, every module is necessary, and a Probe/Online dissociation places the residual difficulty in execution, not perception.

\textbf{Limitations.} Our real-robot evidence is small-scale and on a single Cobot Magic platform, with only $\sim$200 staged always-on streams, so large-scale evidence (especially for the full overload regime, exercised mainly in family D) remains limited. Standing guards are predefined, and the default overload policy uses fixed, dataset-defined priority tiers. Broader real-world data, learned guards and priorities, and richer overload policies are the natural next steps.

\end{document}